\documentclass[conference]{IEEEtran}
\IEEEoverridecommandlockouts
\usepackage{cite}
\usepackage{amsmath,amssymb,amsfonts}
\usepackage{algorithmic}
\usepackage{graphicx}
\usepackage{textcomp}
\usepackage{xcolor}
 \usepackage{booktabs} 
 \usepackage{hyperref} 
\usepackage{colortbl} 
\def\BibTeX{{\rm B\kern-.05em{\sc i\kern-.025em b}\kern-.08em
    T\kern-.1667em\lower.7ex\hbox{E}\kern-.125emX}}
\begin{document}

\title{CU-Mamba: Selective State Space Models with Channel Learning for Image Restoration\\
{}
}

\author{\IEEEauthorblockN{Rui Deng}
\IEEEauthorblockA{\textit{Institute of Computational Mathematical Engineering} \\
\textit{Stanford University}\\
Palo Alto, United States \\
ruideng@stanford.edu}
\and
\IEEEauthorblockN{Tianpei Gu}
\IEEEauthorblockA{\textit{Applied AI Research} \\
\textit{KREA AI}\\
San Francisco, United States \\
g@krea.ai}
}

\maketitle

\begin{abstract}
Reconstructing degraded images is a critical task in image processing. Although CNN and Transformer-based models are prevalent in this field, they exhibit inherent limitations, such as inadequate long-range dependency modeling and high computational costs. To overcome these issues, we introduce the Channel-Aware U-Shaped Mamba (CU-Mamba) model, which incorporates a dual State Space Model (SSM) framework into the U-Net architecture. CU-Mamba employs a Spatial SSM module for global context encoding and a Channel SSM component to preserve channel correlation features, both in linear computational complexity relative to the feature map size. Extensive experimental results validate CU-Mamba's superiority over existing state-of-the-art methods, underscoring the importance of integrating both spatial and channel contexts in image restoration.
\end{abstract}


\section{Introduction}
Image restoration is a fundamental task in digital image processing, aiming to reconstruct high-quality images from those compromised by various degradations such as noise, blur, and rain streaks. Recent advancements have highlighted the effectiveness of Convolutional Neural Networks (CNNs)\cite{dong2023multi,cui2023image,yang2020residual}, and Transformer-based models\cite{zamir2022restormer,wang2022uformer,ke2021musiq,liang2021swinir}, in this domain. CNNs utilize a hierarchical structure that excels in capturing spatial hierarchies within images. Transformer-model are originally designed for natural language
processing, but have shown promising result for visual understanding, such as Vision Transformer\cite{dosovitskiy2020image}. Transformer models employ self-attention mechanisms, making them particularly effective at modeling long-range dependencies. Both approaches have achieved state-of-the-art results in numerous image restoration tasks\cite{yu2024scaling, zhang2021designing, wang2021real}. 

However, both CNNs and Transformer-based models have their limitations. CNNs, despite their effectiveness in local feature extraction, often struggle with capturing long-range dependencies within images due to their limited receptive fields. In contrast, while Transformers mitigate this issue through their global attention modules, they incur quadratic computational costs relative to the size of the feature map. Moreover, Transformers may neglect fine-grained local details that are critical for effective image restoration.

\begin{figure*}
  \centering
  \includegraphics[width=\linewidth]{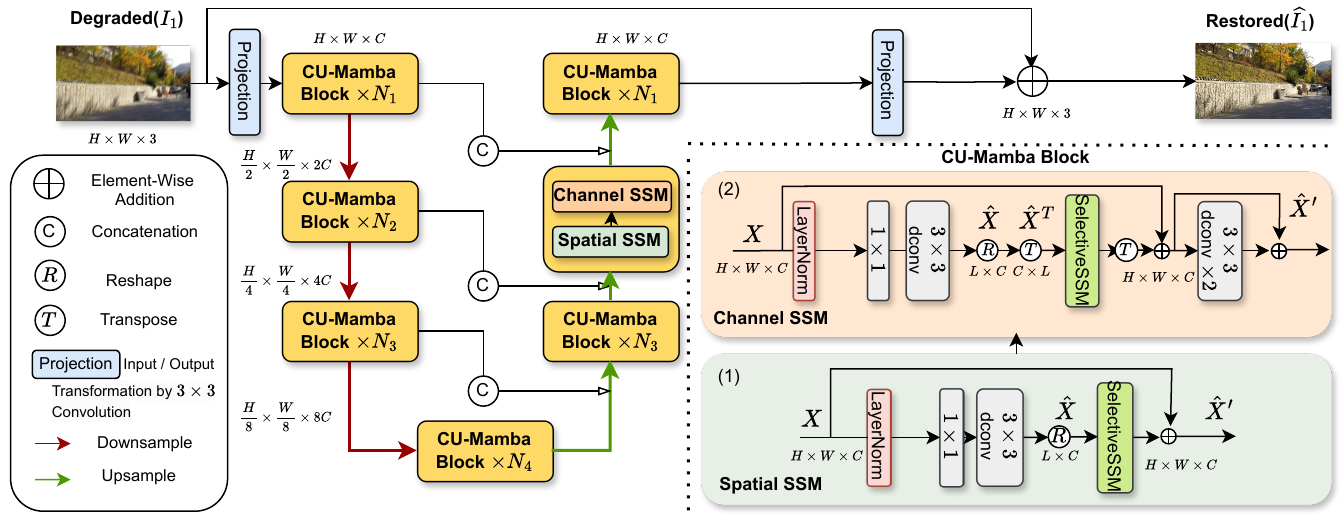}
  \caption{The Overall pipeline of CU-Mamba. Each CU-Mamba Block consists of a Spatial SSM block (as explained in $(1)$) followed by a Channel SSM block (as detailed in $(2)$). The structure of SelectiveSSM block is explained in Fig. \ref{fig:cam_vis}}
  \label{fig:whole_framework}
\end{figure*}

To address these limitations, recent advances have introduced structured state space models (SSMs), particularly the Mamba model\cite{gu2021efficiently, gu2023mamba}, as an effective building block for image recognition networks\cite{nguyen2022s4nd, zhu2024vision}. By efficiently compressing the global context through input-dependent selective SSMs\cite{gu2023mamba}, Mamba maintains the benefits of global receptive fields while operating with linear complexity relative to input tokens.
This approach has demonstrated superior performance in various language and visual tasks, surpassing both CNN and Transformer-based models\cite{gu2023mamba}. However, most visual Mamba models apply SSM blocks independently to each feature channel, leading to a potential loss of information flow across channels\cite{behrouz2024mambamixer}, which is particularly crucial for compressing and reconstructing image details in image restoration.

To solve the above challenges, we propose the Channel-Aware U-Shaped Mamba (CU-Mamba) model for image restoration. On top of the traditional U-Net structure\cite{cho2021rethinking} in image restoration, CU-Mamba attains global receptive fields with Mamba blocks while preserving channel-specific features. We utilize a Spatial State Space Model block within our architecture to effectively capture long-range dependencies in images with linear computational complexity, ensuring a comprehensive understanding of global context. Additionally, we implement a Channel State Space Model component to enhance feature mixing across channels during feature map compression and subsequent upsampling within the U-Net. This dual approach enables the CU-Mamba model to achieve a delicate balance between capturing extensive spatial details and preserving complicated channel-wise correlation, thereby significantly enhancing the quality and accuracy of the restored images.

Overall, the main contributions of this work are:
\begin{itemize}
    \item We introduce the Channel-Aware U-Shaped Mamba (CU-Mamba) model, incorporating a dual State Space Model (SSM) to enrich U-Net with global context and channel-specific features for image restoration.
    \item  We validate the effectiveness of both the Spatial and Channel SSM modules through detailed ablation studies.
    \item Our experiments demonstrate that the CU-Mamba model achieves promising performance on various image restoration datasets, surpassing current SOTA methods while maintaining a lower computational cost.
\end{itemize}.
\section{Related Work}

\textbf{CNN-Based Approaches:} CNN-based models\cite{wang2021real,yang2020residual} have served as a fundamental architecture for image restoration over the past few years. These models demonstrated substantial improvements over traditional techniques\cite{he2010single}, which relied heavily on hand-crafted features and prior knowledge. Among CNN-based models, the U-shaped encoder-decoder networks with skip connection\cite{ronneberger2015u} have demonstrated strong competence in various image restoration tasks due to their hierarchical multi-scale architecture and residual feature representation. 

\textbf{Transformer-Based Approaches:} The inherently local receptive fields in CNNs posed limitations for capturing long-range dependencies. This challenge leads to the adoption of Transformer models\cite{vaswani2017attention, dosovitskiy2020image}, which utilize a global self-attention mechanism to encapsulate long-range interactions across the image. Transformer models are now widely applied in low-level vision tasks such as super-resolution\cite{liang2021swinir}, image denoising\cite{li2023spectral}, deblurring\cite{tsai2022stripformer}, and deraining\cite{chen2023learning}. To reduce the quadratic computation complexity in the attention mechanism, self-attention is performed over local windows\cite{liu2021swin} or channel-dimensions\cite{ding2022davit}. Despite the architectural design, the computational overhead remains high given the intrinsic mechanism of self-attention modules.

\textbf{Visual Structured State Space Models:} Recent innovations include the integration of State Space Models (SSMs)\cite{gu2021efficiently, gu2023mamba} into the image recognition pipeline, as demonstrated by Vision Mamba\cite{zhu2024vision}. SSMs provide a novel method for capturing long-range dependencies with linear computational complexity, thus addressing the computational inefficiencies inherent in Transformers while retaining their global contextual modeling capabilities. U-Mamba\cite{ma2024u} and VM-UNet\cite{ruan2024vm} introduce the Mamba block to the U-net structure, targeting biomedical image segmentation problems. To facilitate the channel-wise information flow, MambaMixer\cite{behrouz2024mambamixer} introduces channel-mixing Mamba blocks to image recognition and time series forecasting. Still, the existing U-shaped Mamba architecture does not integrate channel SSM modules, which are vital for compressing and reconstructing features given the rich context in channel dimensions. In this work, we propose an efficient and effective dual-directional Mamba U-Net that accounts for both the global context and channel correlation during image restoration.

\section{Method}
We aim to develop an effective U-Net that focuses on long-range spatial and channel correlations in image restoration. We propose the CU-Mamba model, which applies the Spatial and Channel SSM blocks to learn global context and channel features with only linear complexity. In this section, we first go through the overall pipeline of our U-Net design, and then we dive into its components by explaining: the selective SSM framework, our Spatial SSM block, and our Channel SSM block. We finally analyze the computation cost of our model to demonstrate its efficiency.

\subsection{Overall Pipeline}
Fig. \ref{fig:whole_framework} demonstrates the overall framework of CU-Mamba. Given a degraded image $I \in \mathbb{R}^{H \times W \times 3}$, it first goes through a $3 \times 3$ convolution to obtain low-level features $X_0 \in \mathbb{R}^{H \times W \times C}$. $X_0$ is then feed through a 4-level symmetrical encoder-decoder U-Net structure to formulate fine-grained, high-quality features. At each level $l$, the encoder consists of $N_l$ CU-Mamba Blocks and a downsampling layer. Specifically, each CU-Mamba Block contains one Spatial SSM block followed by one Channel SSM block, as demonstrated in Fig. \ref{fig:whole_framework} (1) and (2). The downsampling operator hierarchically reduces spatial size and expanding the number of channels, forming the feature map $X_l \in \mathbb{R}^{\frac{H}{2^l} \times \frac{W}{2^l} \times 2^lC}$.

During feature reconstruction, the low-resolution latent feature $X_4 \in \mathbb{R}^{\frac{H}{8} \times \frac{W}{8} \times 8C}$ is passed into the decoder, which consists of $N_l$ CU-Mamba Blocks symmetrical to the encoder and an upsampling layer. The upsampling operator doubles the size of the feature map while reducing the channel length to a half. To facilitate feature reconstruction, we follow \cite{wang2022uformer, zamir2022restormer} to use a skip-connection by concatenating the encoder features with the corresponding decoder features. The final decoded feature is then passed into the output projection block and reshaped back to $R \in \mathbb{R}^{H \times W \times 3}$. We obtain the final restored image: $I' = I + R$. We train the CU-Mamba following the loss function of previous works\cite{wang2022uformer, cho2021rethinking, kong2023efficient}:
\begin{align}
    \mathcal{L}(I', \hat{I}) &= \sqrt{\|I' - \hat{I}\|_2 + \epsilon} + \lambda \| \mathcal{F}(I')- \mathcal{F}(\hat{I})\|_1
\end{align},
where $\hat{I}$ is the ground-truth image, $\mathcal{F}$ is the Fourier transform to the frequency domain. We set $\epsilon=10^{-3}$ and $\lambda=0.1$ in the experiments.

\subsection{Selective SSM Framework}
We provide a simple overview of the selective SSM (Mamba) mechanism\cite{gu2023mamba} employed in our framework. 

Structured state space sequence models (SSMs) map a 1D sequential input $x(t) \in \mathbb{R} \to y(t) \in \mathbb{R}$ through an implicit latent state $h(t) \in \mathbb{R}^N$. A SSM is defined by four parameters $(\Delta, A, B, C)$ with the following operations:
\begin{align}
    h_t &= \bar{A} h_{t-1} + \bar{B} x_t \\
    y_t &= Ch_t
\end{align},
where $(\bar{A}, \bar{B})$ are discrete version of $(A,B)$ from fixed transformations $\bar{A} = f_A(\Delta, A)$ and $\bar{B} = f_B(\Delta, A, B)$. Various discretization rules can be used in a SSM block, and discretization allows efficient parallelizable training through global convolution.

Despite the efficiency brought by discretization, parameters $(\Delta, \bar{A}, \bar{B}, C)$ in SSMs are data-independent and time-invariant, limiting the expressiveness of the hidden state to compress seen context. Selective SSM (or Mamba) introduces data-dependent parameters $(B, C, \Delta)$ that effectively select relevant information in $x_t$: $B_t = \texttt{Linear}_B(x_t), C_t = \texttt{Linear}_C(x_t), \Delta_t = \texttt{SoftPlus}(\texttt{Linear}_{\Delta}(x_t))$.

\begin{figure}[t]
  \centering
  \includegraphics[width=0.55\linewidth]{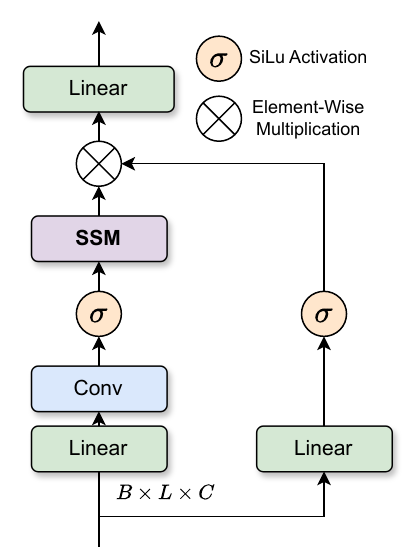}
  \caption{The structure of SelectiveSSM block. On top of the traditional SSM blocks, the selective SSM adds a SiLU activation similar to the Gated MLP\cite{liu2021pay}. This Gated design allows the model to fuse and select information across tokens. On the other hand, the Linear and Conv layers allows the model to learn input-dependent parameters.}
  \label{fig:cam_vis}
\end{figure}

Through hardware-aware optimizations\cite{blelloch1990prefix, smith2022simplified}, selective SSM consumes linear computation and memory complexity w.r.t. sequence lengths, while effectively compressing relevant context over the global input sequence. The optimized selective SSM (Mamba) architecture\cite{gu2023mamba} is shown in Fig. \ref{fig:cam_vis}.

\subsection{Global Learning Block: Spatial SSM}
The success of the transformer architectures\cite{zamir2022restormer, tsai2022stripformer} shows that integrating the global context through the hierarchical structure of the U-Net is crucial for high-quality image restoration. However, such global receptive field comes at the cost of quadratic computational complexity\cite{vaswani2017attention}. We thus design a global learning block that effectively compresses the long-range context with the selective SSM framework, which only requires linear computational complexity.

Given a layer-normalized input tensor $X \in \mathbb{R}^{H \times W \times C}$, we first apply $1 \times 1$ convolutions to gather context across different channels at the pixel level, and then use $3 \times 3$ depth-wise convolutions to capture the spatial context through channels. We then flatten the feature map into $\hat{X} \in \mathbb{R}^{L \times C}$, where $L = H \times W$, to construct a sequence of feature patches. We encode the global context of $\hat{X}$ through:
\begin{align}
    \hat{X'} &= \texttt{SelectiveSSM}(\hat{X})
\end{align},
where the $\texttt{SelectiveSSM}$ block is demonstrated and explained in Fig. \ref{fig:cam_vis}. We can interpret this operation as linearly scanning the feature map of tensor $X$ from the upper-left to the bottom-right corner, where each pixel in the map learns its hidden representation from all previously seen context. The final representation $\hat{X'}$ is reshaped to $\hat{X'} \in \mathbb{R}^{H \times W \times C}$, and it encodes the long-range dependency within its $H \times W$ dimensions.

\begin{figure*}
  \centering
  \includegraphics[width=0.7\linewidth]{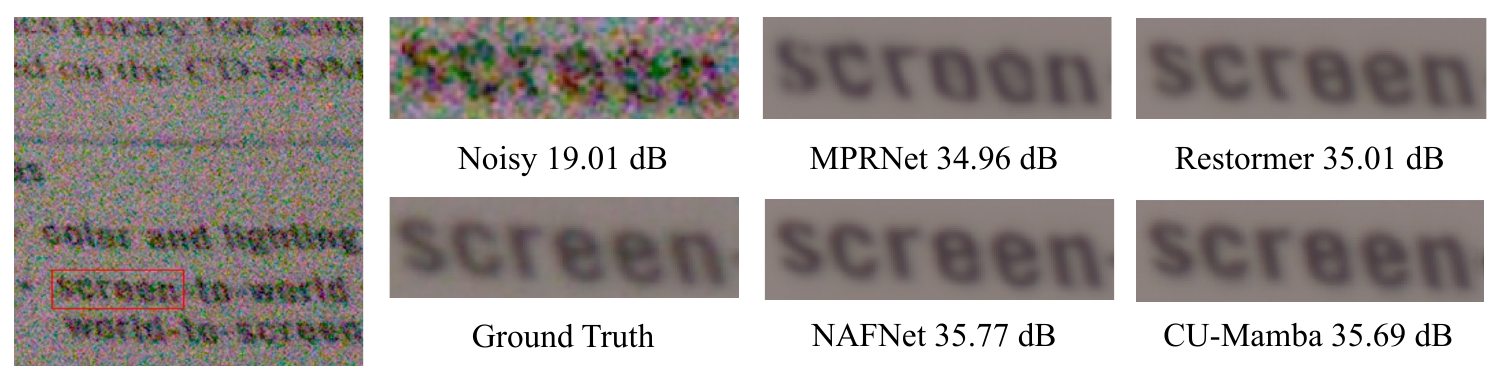}
  \caption{The visualization of image denosing result in SIDD\cite{abdelhamed2018high} dataset.}
  \label{fig:noise}
\end{figure*}

\subsection{Channel Learning Block: Channel SSM}
In U-Net architectures, the channel-wise features in the downsampling and upsampling path is essential for compressing and reconstructing image context and structures. One problem of the existing Mamba-based U-Net is that while the global context is captured by scanning image feature map, the channel information is usually overlooked.\cite{behrouz2024mambamixer} To learn the dependencies of features across channels, we introduce the selective SSM mechanism to the channel dimension.

Similar to the Sptial SSM block, given a layer-normalized input tensor $X \in \mathbb{R}^{H \times W \times C}$, we use $1 \times 1$ convolutions followed by $3 \times 3$ depth-wise convolutions to preprocess the local context. We then transpose $X$ to $X^T \in \mathbb{R}^{C \times H \times W}$ and flatten to $\hat{X^T} \in \mathbb{R}^{C \times L}$. This can be viewed as using flattened feature pixels as the channel representation. We then apply the selective SSM by:
\begin{align}
    \hat{X^{T'}} &= \texttt{SelectiveSSM}(\hat{X^T})
\end{align}
This operation effectively mixes and memorizes the channel-wise feature by scanning the channel map from top to bottom. The final feature $\hat{X^{T'}}$ is reshaped and transposed back to $\hat{X^{*}} \in \mathbb{R}^{H \times W \times C}$. It is then passed to 2 blocks of $3 \times 3$ depth-wise convolutions with $\texttt{LeakyReLU}$ activation to smooth the local representations.

\subsection{Computational Complexity of CU-Mamba}
We follow the complexity analysis from \cite{gu2023mamba} and \cite{behrouz2024mambamixer}. Denote the batch size as $B$, the input sequence length as $L$ (Here, $L = H \times W$), the channel dimension as $C$, and the expansion factor as $E$ ($E=2$ throughout our implementation). With the efficient parallel scan algorithm\cite{smith2022simplified}, the computational complexity is $\mathcal{O}(BLE + EC)$\cite{gu2023mamba} for the spatial SSM block and $\mathcal{O}(BCE + EL)$ for the channel SSM block. The total complexity is thus $\mathcal{O}(BE(L + C))$, which is linear in the sequence length and channel dimensions.


\begin{table}[]
\begin{center}
\caption{Real Noise Removal Experimental Results.}
\label{tab:table1}
\resizebox{0.8\linewidth}{!}{
\begin{tabular}{c|cc|cc}
    \toprule
         & \multicolumn{2}{c|}{\textbf{SIDD}\cite{abdelhamed2018high}} & \multicolumn{2}{c}{\textbf{DND}\cite{plotz2017benchmarking}} \\
Model    & PSNR        & \cellcolor{gray!20} SSIM       & PSNR       & \cellcolor{gray!20} SSIM       \\
\midrule
BM3D \cite{dabov2007image}        &     25.65        &     \cellcolor{gray!20}  0.685    &    34.51        &    \cellcolor{gray!20}   0.851     \\
RIDNet \cite{anwar2019real}       &     38.71        &       \cellcolor{gray!20}  0.914  &     39.26       &     \cellcolor{gray!20}    0.953   \\
DAGL \cite{mou2021dynamic}        &        38.94     &       \cellcolor{gray!20}  0.953   &     39.77       &     \cellcolor{gray!20}  0.956     \\
DANet+  \cite{yue2020dual}      &     39.47        &       \cellcolor{gray!20}  0.918   &      39.59      &     \cellcolor{gray!20}    0.955   \\
MPRNet   \cite{zamir2021multi}     &    39.71         &       \cellcolor{gray!20} 0.958    &      39.80      &     \cellcolor{gray!20}   0.954    \\
Uformer  \cite{wang2022uformer}      &   39.77          &       \cellcolor{gray!20}  0.959   &     39.96       &     \cellcolor{gray!20}  0.956     \\
HINet  \cite{chen2021hinet}      &   39.99          &       \cellcolor{gray!20}  0.958   &     -       &     \cellcolor{gray!20} -     \\
Restormer   \cite{zamir2022restormer}     &   40.02          &       \cellcolor{gray!20} \underline{0.960}    &       \underline{40.03}    &     \cellcolor{gray!20}    \underline{0.956}   \\
NAFNet   \cite{chen2022simple}     &    \textbf{40.30}         &      \cellcolor{gray!20} \textbf{0.962}    &       -     &     \cellcolor{gray!20}     -  \\

\midrule
CU-Mamba &   \underline{40.22}          &     \cellcolor{gray!20}  \textbf{0.962}     &     \textbf{40.34}       & \cellcolor{gray!20} \textbf{0.960} \\
\bottomrule
\end{tabular}
}
\end{center}
\end{table}

\begin{table*}[ht]
\begin{center}
\caption{Image Motion Deblurring Experimental Results.}
\label{tab:table2}
\resizebox{0.7\linewidth}{!}{
\begin{tabular}{c|cc|cc|cc|cc}
    \toprule
         & \multicolumn{2}{c|}{\textbf{GoPro}\cite{nah2017deep}}       & \multicolumn{2}{c|}{\textbf{HIDE}\cite{shen2019human}}        & \multicolumn{2}{c|}{\textbf{RealBlur-R}\cite{rim2020real}}  & \multicolumn{2}{c}{\textbf{RealBlur-J}\cite{rim2020real}}  \\
Model    & PSNR & \multicolumn{1}{l|}{\cellcolor{gray!20}  SSIM} & PSNR & \multicolumn{1}{l|}{\cellcolor{gray!20}  SSIM} & PSNR & \multicolumn{1}{l|}{\cellcolor{gray!20}  SSIM} & PSNR           & \cellcolor{gray!20}  SSIM           \\
    \midrule
 Xu \textit{et al.} \cite{xu2013unnatural}     & 21.00     &  \cellcolor{gray!20}  0.741                       &    -  &     \cellcolor{gray!20}  -                   &    34.46   &      \cellcolor{gray!20}  0.937                     &          27.14      &    \cellcolor{gray!20}  0.830             \\
  Nah \textit{et al.} \cite{nah2017deep}    &   29.08   &  \cellcolor{gray!20}  0.914                       &  25.73    &     \cellcolor{gray!20}  0.874                      &  32.51    &      \cellcolor{gray!20}  0.841                     &            27.87    &    \cellcolor{gray!20}  0.827             \\
  DeblurGAN-v2 \cite{kupyn2019deblurgan}     & 29.55     &  \cellcolor{gray!20}  0.934                       &    26.61  &     \cellcolor{gray!20}  0.875                      &   35.26   &      \cellcolor{gray!20}  0.944                     &            28.70    &    \cellcolor{gray!20}  0.866             \\
  DMPHN \cite{zhang2019deep}    &   31.20   &  \cellcolor{gray!20}   0.940                       &    29.09  &     \cellcolor{gray!20}  0.924                     &   35.70   &      \cellcolor{gray!20}  0.948                     &        28.42      &    \cellcolor{gray!20}  0.860             \\
  SPAIR \cite{purohit2021spatially}    &   32.06   &  \cellcolor{gray!20}  0.953                       &    30.29  &     \cellcolor{gray!20}  0.931                      & -     &      \cellcolor{gray!20}  -                     &        28.81       &    \cellcolor{gray!20}  0.875             \\
  MPRNet \cite{zamir2021multi}     &   32.66   &  \cellcolor{gray!20}  0.959                       &    30.96  &     \cellcolor{gray!20}  0.939                     &  35.99    &      \cellcolor{gray!20}  0.952                     &        28.70       &    \cellcolor{gray!20}  0.873             \\   
  Restormer \cite{zamir2022restormer}     &   32.92   &  \cellcolor{gray!20}  0.961                  &   31.22   &     \cellcolor{gray!20}  0.942                      &  36.19   &      \cellcolor{gray!20}  0.957                     &         28.96      &    \cellcolor{gray!20}  0.879              \\  
  Uformer \cite{wang2022uformer}    &   32.97   &  \cellcolor{gray!20}  0.967                       &  30.83    &     \cellcolor{gray!20}  0.952                      &   36.22   &      \cellcolor{gray!20}  0.957                     &      29.06        &    \cellcolor{gray!20}  0.884             \\      
  Stripformer \cite{tsai2022stripformer}     &    33.08 &  \cellcolor{gray!20}  0.962                       &  31.03    &     \cellcolor{gray!20}  0.940                      &    39.84  &      \cellcolor{gray!20}  0.974                     &          32.48    &    \cellcolor{gray!20}  \underline{0.929}             \\      
  FSNet \cite{cui2023image}    &    33.29  &  \cellcolor{gray!20}  \underline{0.963}                       &  31.05    &     \cellcolor{gray!20}  0.941                      &   35.84   &      \cellcolor{gray!20}  0.952                     &         -      &    \cellcolor{gray!20}  -             \\      
  MRLPFNet \cite{dong2023multi}    &    \underline{33.50}  &  \cellcolor{gray!20} \textbf{0.965}                        &   \textbf{31.63}   &     \cellcolor{gray!20}   \textbf{0.946}                     &  \underline{40.92}    &      \cellcolor{gray!20}  \underline{0.975}                   &         \underline{33.19}      &    \cellcolor{gray!20}  \textbf{0.936}             \\      

    \midrule
CU-Mamba &   \textbf{33.53}   &      \cellcolor{gray!20}  \textbf{0.965}                     &  \underline{31.47}    &     \cellcolor{gray!20} \underline{0.944}                      &  \textbf{41.01}    &       \cellcolor{gray!20}  \textbf{0.976}                   &   \textbf{33.21}             &   \cellcolor{gray!20}  \textbf{0.936}          \\
    \bottomrule
\end{tabular}
}
\end{center}
\end{table*}

\section{Experiments}


\begin{figure*}
  \centering
  \includegraphics[width=\linewidth]{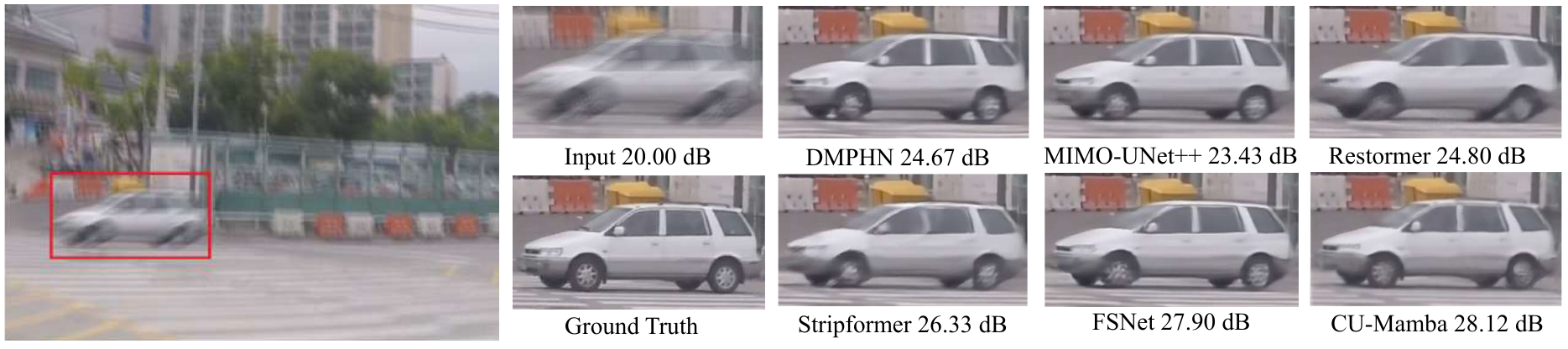}
  \caption{The visualization of image motion deblurring result in GoPro\cite{nah2017deep} dataset.}
  \label{fig:demo_1}
\end{figure*}

We first explain the details of our experimental settings. Then, we demonstrate the strong performance of CU-Mamba through extensive experiments in image denoising and image deblurring. We finally conduct ablation studies to validate the effectiveness of each module in the CU-Mamba model.
\subsection{Experimental Settings}
\textbf{Parameter Settings:} Following the previous training setting\cite{wang2022uformer}, we preprocess the training samples by randomly flipping the image horizontally and rotating the images by $90^\circ$, $180^\circ$, or $270^\circ$. During training, we use the AdamW optimizer\cite{loshchilov2017decoupled} with the momentum $\beta_1=0.9$ and $\beta_2=0.999$. To stabilize the training process, we set the initial learning rate to $5e^{-5}$ and gradually decrease to $1e^{-6}$ with the cosine annealing strategy\cite{loshchilov2016sgdr}. The channel width $C$ passed to the initial SSM block is set to 32.

\textbf{Evaluation Metrics:} To evaluate the restoration quality, we use the PSNR (Peak Signal-to-Noise Ratio) and SSIM (Structural Similarity Index)\cite{wang2004image} metrics following previous works. PSNR measures the quality of a reconstructed image by calculating the ratio between the maximum possible signal power and the power of corrupting noise, while SSIM compares the structural similarity between the reconstructed and the original image. We calculate these metrics under the RGB color space. 

\subsection{Image Denoising Results}
We evaluated the performance of our method for real-world noise removal using the SIDD\cite{abdelhamed2018high} and DND\cite{plotz2017benchmarking} datasets. For evaluation of the DND dataset, we follow the common strategy of previous works\cite{wang2022uformer}: train our model using the SIDD dataset and test our model in DND's online server. Table \ref{tab:table1} presents a comparative analysis of our proposed method with state-of-the-art approaches on the SIDD and DND datasets. Notably, our approach outperforms existing CNN-based methods (RIDNet\cite{anwar2019real}, MPRNet\cite{zamir2021multi}, HINet\cite{chen2021hinet}) and Transformer-based methods (Uformer\cite{wang2022uformer}, Restormer\cite{zamir2022restormer}), showcasing its efficacy in handling real-world noise. Fig. \ref{fig:noise} showcases the qualitative results of our method compared to ground truth images on the SIDD dataset. We can observe that our method reconstruct the noisy image with more exact details similar to the ground-truth image.

\subsection{Image Deblurring Results}
We test the motion blur removal performance of CU-Mamba in four datasets. Following the previous methods\cite{zamir2021multi}, we train our model using the training set of the GoPro dataset\cite{nah2017deep}, and then we test our model on two synthetic (Test set of GoPro\cite{nah2017deep} and HIDE\cite{shen2019human}) and two real-world datasets (RealBlur-R and RealBlur-J\cite{rim2020real}). As demonstrated in Table \ref{tab:table2}, our method outperforms the current state-of-the-art MRLPFNet\cite{dong2023multi}, achieving a 0.09 dB improvement for real-world image deblurring on the RealBlur-R dataset. We also test and compare the inference time of our model with previous works in table \ref{tab:table4}. Our model exhibits a $4\times$ faster inference speed comparing to the transformer-based model Restormer\cite{zamir2022restormer}, additionally achieving a 0.87 dB performance gain in PSNR. This proves the efficiency of linear complexity in our selective SSM blocks over the quadratic cost of self-attention modules. Moreover, through qualitative comparisons with other existing methods in Fig. \ref{fig:demo_1}, we highlight the effectiveness of CU-Mamba in producing more realistic deblurred images that aligns closely to the ground-truth label.

\begin{table}[]
\begin{center}
\caption{The inference time comparison with existing models using GoPro test dataset\cite{nah2017deep}. FLOPs is calculated with input image of size $256 \times 256$.}
\label{tab:table4}
\resizebox{0.8\linewidth}{!}{
\begin{tabular}{cccc}
    \toprule
          & Time (s) & FLOPs (G) & PSNR \\
              \midrule
DBGAN\cite{zhang2020deblurring}   &    1.447      &  759.85    &   31.10   \\
DMPHN\cite{zhang2019deep}   &    0.405      &   -   &  31.20    \\
MPRNet\cite{zamir2021multi}   &   1.148       &   777.01   &  32.66    \\
Restormer\cite{zamir2022restormer} &   1.218       &  140.99    &   32.92   \\
FSNet\cite{cui2023image} &   \underline{0.362}       &  \underline{111.14}    &   \underline{33.29}   \\

    \midrule

CU-Mamba  &   \textbf{0.305}       &  \textbf{43.10}    &   \textbf{33.53}  \\
    \bottomrule

\end{tabular}
}
\end{center}
\end{table}

\begin{table}[]
\begin{center}
\caption{Ablation Study of CU-Mamba Components. PSNR and SSIM is calculated using GoPro test dataset\cite{nah2017deep}.}
\label{tab:table3}
\resizebox{\linewidth}{!}{
\begin{tabular}{c|c|cc}
    \toprule
Architecture                                & \# Params (M) & PSNR & SSIM  \\
    \midrule

UNet with Resblocks\cite{cho2021rethinking} &    16.1       &   32.45   &      0.957  \\
    \midrule
\checkmark Spatial SSM                                 &   15.6        &   \underline{33.31}   &   \underline{0.963}    \\
\checkmark Channel SSM                                 &    16.3       &  33.07    &   0.962    \\
    \midrule
\checkmark \checkmark
 Spatial + Channel SSM                       &    19.7       &   \textbf{33.53}   &   \textbf{0.965}    \\
    \bottomrule
\end{tabular}
}
\end{center}
\end{table}

\subsection{Ablation Studies}
We conduct ablation studies for the spatial and channel SSM modules of CU-Mamba to analyze their impacts. Table \ref{tab:table3} shows the number of model parameter, SSIM, and PSNR of each ablation setting in the GoPro test dataset\cite{nah2017deep}. The spatial SSM block, when applied independently, exhibited a more effective improvement in model performance compared to the channel SSM block alone. This suggests that the long-range dependency plays a crucial role in image restoration, and the channel SSM alone is not able to encode global information. The channel SSM, on the other hand, facilitates the information mixing across the channel direction, thus it also shows 0.62 dB increase in PSNR over the UNet baseline. Still, when we apply both the spatial and channel blocks, CU-Mamba demonstrates the best performance of 33.53 dB in PSNR. This highlights the effectiveness of integrating both types of selective SSM blocks to fully exploit the model's capacity of learning complex, hierarchical image representations.



\section{Conclusion}
In conclusion, we introduce the Channel-Aware U-Shaped Mamba (CU-Mamba) model, which improves image restoration by combining a U-Net framework with dual-direction selective State Space Models to better understand and reconstruct images. Extensive experiments show that CU-Mamba performs well compared to existing methods, demonstrating the efficiency and effectiveness of our approach. This work offers a new perspective on the U-Net architecture in image restoration, showcasing the importance of both the spatial and channel context in feature reconstruction and opening a promising direction for further research and practical applications.



{
\bibliographystyle{IEEEtran}
\bibliography{egbib}

\begin{thebibliography}{10}
\providecommand{\url}[1]{#1}
\csname url@samestyle\endcsname
\providecommand{\newblock}{\relax}
\providecommand{\bibinfo}[2]{#2}
\providecommand{\BIBentrySTDinterwordspacing}{\spaceskip=0pt\relax}
\providecommand{\BIBentryALTinterwordstretchfactor}{4}
\providecommand{\BIBentryALTinterwordspacing}{\spaceskip=\fontdimen2\font plus
\BIBentryALTinterwordstretchfactor\fontdimen3\font minus \fontdimen4\font\relax}
\providecommand{\BIBforeignlanguage}[2]{{%
\expandafter\ifx\csname l@#1\endcsname\relax
\typeout{** WARNING: IEEEtran.bst: No hyphenation pattern has been}%
\typeout{** loaded for the language `#1'. Using the pattern for}%
\typeout{** the default language instead.}%
\else
\language=\csname l@#1\endcsname
\fi
#2}}
\providecommand{\BIBdecl}{\relax}
\BIBdecl

\bibitem{dong2023multi}
J.~Dong, J.~Pan, Z.~Yang, and J.~Tang, ``Multi-scale residual low-pass filter network for image deblurring,'' in \emph{Proceedings of the IEEE/CVF International Conference on Computer Vision}, 2023, pp. 12\,345--12\,354.

\bibitem{cui2023image}
Y.~Cui, W.~Ren, X.~Cao, and A.~Knoll, ``Image restoration via frequency selection,'' \emph{IEEE Transactions on Pattern Analysis and Machine Intelligence}, 2023.

\bibitem{yang2020residual}
Q.~Yang, Y.~Liu, J.~Tang, and T.~Ku, ``Residual and dense unet for under-display camera restoration,'' in \emph{European Conference on Computer Vision}.\hskip 1em plus 0.5em minus 0.4em\relax Springer, 2020, pp. 398--408.

\bibitem{zamir2022restormer}
S.~W. Zamir, A.~Arora, S.~Khan, M.~Hayat, F.~S. Khan, and M.-H. Yang, ``Restormer: Efficient transformer for high-resolution image restoration,'' in \emph{Proceedings of the IEEE/CVF conference on computer vision and pattern recognition}, 2022, pp. 5728--5739.

\bibitem{wang2022uformer}
Z.~Wang, X.~Cun, J.~Bao, W.~Zhou, J.~Liu, and H.~Li, ``Uformer: A general u-shaped transformer for image restoration,'' in \emph{Proceedings of the IEEE/CVF conference on computer vision and pattern recognition}, 2022, pp. 17\,683--17\,693.

\bibitem{ke2021musiq}
J.~Ke, Q.~Wang, Y.~Wang, P.~Milanfar, and F.~Yang, ``Musiq: Multi-scale image quality transformer,'' in \emph{Proceedings of the IEEE/CVF international conference on computer vision}, 2021, pp. 5148--5157.

\bibitem{liang2021swinir}
J.~Liang, J.~Cao, G.~Sun, K.~Zhang, L.~Van~Gool, and R.~Timofte, ``Swinir: Image restoration using swin transformer,'' in \emph{Proceedings of the IEEE/CVF international conference on computer vision}, 2021, pp. 1833--1844.

\bibitem{dosovitskiy2020image}
A.~Dosovitskiy, L.~Beyer, A.~Kolesnikov, D.~Weissenborn, X.~Zhai, T.~Unterthiner, M.~Dehghani, M.~Minderer, G.~Heigold, S.~Gelly \emph{et~al.}, ``An image is worth 16x16 words: Transformers for image recognition at scale,'' 2021.

\bibitem{yu2024scaling}
F.~Yu, J.~Gu, Z.~Li, J.~Hu, X.~Kong, X.~Wang, J.~He, Y.~Qiao, and C.~Dong, ``Scaling up to excellence: Practicing model scaling for photo-realistic image restoration in the wild,'' 2024.

\bibitem{zhang2021designing}
K.~Zhang, J.~Liang, L.~Van~Gool, and R.~Timofte, ``Designing a practical degradation model for deep blind image super-resolution,'' in \emph{Proceedings of the IEEE/CVF International Conference on Computer Vision}, 2021, pp. 4791--4800.

\bibitem{wang2021real}
X.~Wang, L.~Xie, C.~Dong, and Y.~Shan, ``Real-esrgan: Training real-world blind super-resolution with pure synthetic data,'' in \emph{Proceedings of the IEEE/CVF international conference on computer vision}, 2021, pp. 1905--1914.

\bibitem{gu2021efficiently}
A.~Gu, K.~Goel, and C.~R{\'e}, ``Efficiently modeling long sequences with structured state spaces,'' 2022.

\bibitem{gu2023mamba}
A.~Gu and T.~Dao, ``Mamba: Linear-time sequence modeling with selective state spaces,'' \emph{arXiv preprint arXiv:2312.00752}, 2023.

\bibitem{nguyen2022s4nd}
E.~Nguyen, K.~Goel, A.~Gu, G.~Downs, P.~Shah, T.~Dao, S.~Baccus, and C.~R{\'e}, ``S4nd: Modeling images and videos as multidimensional signals with state spaces,'' \emph{Advances in neural information processing systems}, vol.~35, pp. 2846--2861, 2022.

\bibitem{zhu2024vision}
L.~Zhu, B.~Liao, Q.~Zhang, X.~Wang, W.~Liu, and X.~Wang, ``Vision mamba: Efficient visual representation learning with bidirectional state space model,'' \emph{arXiv preprint arXiv:2401.09417}, 2024.

\bibitem{behrouz2024mambamixer}
A.~Behrouz, M.~Santacatterina, and R.~Zabih, ``Mambamixer: Efficient selective state space models with dual token and channel selection,'' \emph{arXiv preprint arXiv:2403.19888}, 2024.

\bibitem{cho2021rethinking}
S.-J. Cho, S.-W. Ji, J.-P. Hong, S.-W. Jung, and S.-J. Ko, ``Rethinking coarse-to-fine approach in single image deblurring,'' in \emph{Proceedings of the IEEE/CVF international conference on computer vision}, 2021, pp. 4641--4650.

\bibitem{he2010single}
K.~He, J.~Sun, and X.~Tang, ``Single image haze removal using dark channel prior,'' \emph{IEEE transactions on pattern analysis and machine intelligence}, vol.~33, no.~12, pp. 2341--2353, 2010.

\bibitem{ronneberger2015u}
O.~Ronneberger, P.~Fischer, and T.~Brox, ``U-net: Convolutional networks for biomedical image segmentation,'' in \emph{Medical image computing and computer-assisted intervention--MICCAI 2015: 18th international conference, Munich, Germany, October 5-9, 2015, proceedings, part III 18}.\hskip 1em plus 0.5em minus 0.4em\relax Springer, 2015, pp. 234--241.

\bibitem{vaswani2017attention}
A.~Vaswani, N.~Shazeer, N.~Parmar, J.~Uszkoreit, L.~Jones, A.~N. Gomez, {\L}.~Kaiser, and I.~Polosukhin, ``Attention is all you need,'' \emph{Advances in neural information processing systems}, vol.~30, 2017.

\bibitem{li2023spectral}
M.~Li, J.~Liu, Y.~Fu, Y.~Zhang, and D.~Dou, ``Spectral enhanced rectangle transformer for hyperspectral image denoising,'' in \emph{Proceedings of the IEEE/CVF Conference on Computer Vision and Pattern Recognition}, 2023, pp. 5805--5814.

\bibitem{tsai2022stripformer}
F.-J. Tsai, Y.-T. Peng, Y.-Y. Lin, C.-C. Tsai, and C.-W. Lin, ``Stripformer: Strip transformer for fast image deblurring,'' in \emph{European Conference on Computer Vision}.\hskip 1em plus 0.5em minus 0.4em\relax Springer, 2022, pp. 146--162.

\bibitem{chen2023learning}
X.~Chen, H.~Li, M.~Li, and J.~Pan, ``Learning a sparse transformer network for effective image deraining,'' in \emph{Proceedings of the IEEE/CVF Conference on Computer Vision and Pattern Recognition}, 2023, pp. 5896--5905.

\bibitem{liu2021swin}
Z.~Liu, Y.~Lin, Y.~Cao, H.~Hu, Y.~Wei, Z.~Zhang, S.~Lin, and B.~Guo, ``Swin transformer: Hierarchical vision transformer using shifted windows,'' in \emph{Proceedings of the IEEE/CVF international conference on computer vision}, 2021, pp. 10\,012--10\,022.

\bibitem{ding2022davit}
M.~Ding, B.~Xiao, N.~Codella, P.~Luo, J.~Wang, and L.~Yuan, ``Davit: Dual attention vision transformers,'' in \emph{European conference on computer vision}.\hskip 1em plus 0.5em minus 0.4em\relax Springer, 2022, pp. 74--92.

\bibitem{ma2024u}
J.~Ma, F.~Li, and B.~Wang, ``U-mamba: Enhancing long-range dependency for biomedical image segmentation,'' \emph{arXiv preprint arXiv:2401.04722}, 2024.

\bibitem{ruan2024vm}
J.~Ruan and S.~Xiang, ``Vm-unet: Vision mamba unet for medical image segmentation,'' \emph{arXiv preprint arXiv:2402.02491}, 2024.

\bibitem{kong2023efficient}
L.~Kong, J.~Dong, J.~Ge, M.~Li, and J.~Pan, ``Efficient frequency domain-based transformers for high-quality image deblurring,'' in \emph{Proceedings of the IEEE/CVF Conference on Computer Vision and Pattern Recognition}, 2023, pp. 5886--5895.

\bibitem{blelloch1990prefix}
G.~E. Blelloch, ``Prefix sums and their applications,'' 1990.

\bibitem{smith2022simplified}
J.~T. Smith, A.~Warrington, and S.~W. Linderman, ``Simplified state space layers for sequence modeling,'' \emph{arXiv preprint arXiv:2208.04933}, 2022.

\bibitem{liu2021pay}
H.~Liu, Z.~Dai, D.~So, and Q.~V. Le, ``Pay attention to mlps,'' \emph{Advances in neural information processing systems}, vol.~34, pp. 9204--9215, 2021.

\bibitem{abdelhamed2018high}
A.~Abdelhamed, S.~Lin, and M.~S. Brown, ``A high-quality denoising dataset for smartphone cameras,'' in \emph{Proceedings of the IEEE conference on computer vision and pattern recognition}, 2018, pp. 1692--1700.

\bibitem{plotz2017benchmarking}
T.~Plotz and S.~Roth, ``Benchmarking denoising algorithms with real photographs,'' in \emph{Proceedings of the IEEE conference on computer vision and pattern recognition}, 2017, pp. 1586--1595.

\bibitem{dabov2007image}
K.~Dabov, A.~Foi, V.~Katkovnik, and K.~Egiazarian, ``Image denoising by sparse 3-d transform-domain collaborative filtering,'' \emph{IEEE Transactions on image processing}, vol.~16, no.~8, pp. 2080--2095, 2007.

\bibitem{anwar2019real}
S.~Anwar and N.~Barnes, ``Real image denoising with feature attention,'' in \emph{Proceedings of the IEEE/CVF international conference on computer vision}, 2019, pp. 3155--3164.

\bibitem{mou2021dynamic}
C.~Mou, J.~Zhang, and Z.~Wu, ``Dynamic attentive graph learning for image restoration,'' in \emph{Proceedings of the IEEE/CVF international conference on computer vision}, 2021, pp. 4328--4337.

\bibitem{yue2020dual}
Z.~Yue, Q.~Zhao, L.~Zhang, and D.~Meng, ``Dual adversarial network: Toward real-world noise removal and noise generation,'' in \emph{Computer Vision--ECCV 2020: 16th European Conference, Glasgow, UK, August 23--28, 2020, Proceedings, Part X 16}.\hskip 1em plus 0.5em minus 0.4em\relax Springer, 2020, pp. 41--58.

\bibitem{zamir2021multi}
S.~W. Zamir, A.~Arora, S.~Khan, M.~Hayat, F.~S. Khan, M.-H. Yang, and L.~Shao, ``Multi-stage progressive image restoration,'' in \emph{Proceedings of the IEEE/CVF conference on computer vision and pattern recognition}, 2021, pp. 14\,821--14\,831.

\bibitem{chen2021hinet}
L.~Chen, X.~Lu, J.~Zhang, X.~Chu, and C.~Chen, ``Hinet: Half instance normalization network for image restoration,'' in \emph{Proceedings of the IEEE/CVF Conference on Computer Vision and Pattern Recognition}, 2021, pp. 182--192.

\bibitem{chen2022simple}
L.~Chen, X.~Chu, X.~Zhang, and J.~Sun, ``Simple baselines for image restoration,'' in \emph{European conference on computer vision}.\hskip 1em plus 0.5em minus 0.4em\relax Springer, 2022, pp. 17--33.

\bibitem{nah2017deep}
S.~Nah, T.~Hyun~Kim, and K.~Mu~Lee, ``Deep multi-scale convolutional neural network for dynamic scene deblurring,'' in \emph{Proceedings of the IEEE conference on computer vision and pattern recognition}, 2017, pp. 3883--3891.

\bibitem{shen2019human}
Z.~Shen, W.~Wang, X.~Lu, J.~Shen, H.~Ling, T.~Xu, and L.~Shao, ``Human-aware motion deblurring,'' in \emph{Proceedings of the IEEE/CVF International Conference on Computer Vision}, 2019, pp. 5572--5581.

\bibitem{rim2020real}
J.~Rim, H.~Lee, J.~Won, and S.~Cho, ``Real-world blur dataset for learning and benchmarking deblurring algorithms,'' in \emph{Computer Vision--ECCV 2020: 16th European Conference, Glasgow, UK, August 23--28, 2020, Proceedings, Part XXV 16}.\hskip 1em plus 0.5em minus 0.4em\relax Springer, 2020, pp. 184--201.

\bibitem{xu2013unnatural}
L.~Xu, S.~Zheng, and J.~Jia, ``Unnatural l0 sparse representation for natural image deblurring,'' in \emph{Proceedings of the IEEE conference on computer vision and pattern recognition}, 2013, pp. 1107--1114.

\bibitem{kupyn2019deblurgan}
O.~Kupyn, T.~Martyniuk, J.~Wu, and Z.~Wang, ``Deblurgan-v2: Deblurring (orders-of-magnitude) faster and better,'' in \emph{Proceedings of the IEEE/CVF international conference on computer vision}, 2019, pp. 8878--8887.

\bibitem{zhang2019deep}
H.~Zhang, Y.~Dai, H.~Li, and P.~Koniusz, ``Deep stacked hierarchical multi-patch network for image deblurring,'' in \emph{Proceedings of the IEEE/CVF conference on computer vision and pattern recognition}, 2019, pp. 5978--5986.

\bibitem{purohit2021spatially}
K.~Purohit, M.~Suin, A.~Rajagopalan, and V.~N. Boddeti, ``Spatially-adaptive image restoration using distortion-guided networks,'' in \emph{Proceedings of the IEEE/CVF international conference on computer vision}, 2021, pp. 2309--2319.

\bibitem{loshchilov2017decoupled}
I.~Loshchilov and F.~Hutter, ``Decoupled weight decay regularization,'' \emph{arXiv preprint arXiv:1711.05101}, 2017.

\bibitem{loshchilov2016sgdr}
------, ``Sgdr: Stochastic gradient descent with warm restarts,'' \emph{arXiv preprint arXiv:1608.03983}, 2016.

\bibitem{wang2004image}
Z.~Wang, A.~C. Bovik, H.~R. Sheikh, and E.~P. Simoncelli, ``Image quality assessment: from error visibility to structural similarity,'' \emph{IEEE transactions on image processing}, vol.~13, no.~4, pp. 600--612, 2004.

\bibitem{zhang2020deblurring}
K.~Zhang, W.~Luo, Y.~Zhong, L.~Ma, B.~Stenger, W.~Liu, and H.~Li, ``Deblurring by realistic blurring,'' in \emph{Proceedings of the IEEE/CVF conference on computer vision and pattern recognition}, 2020, pp. 2737--2746.

\end{thebibliography}
}

\end{document}